\documentclass[letterpaper]{article} 
\usepackage{aaai2027}  
\usepackage[hyphens]{url}  
\usepackage{graphicx} 
\usepackage{natbib}  
\usepackage{caption} 
\usepackage{algorithm}
\usepackage{algorithmic}

\usepackage{newfloat}
\usepackage{listings}
\DeclareCaptionStyle{ruled}{labelfont=normalfont,labelsep=colon,strut=off} 
\floatstyle{ruled}
\newfloat{listing}{tb}{lst}{}
\floatname{listing}{Listing}

\usepackage{booktabs}

\usepackage{adjustbox}
\usepackage{pifont}
\usepackage{xcolor}
\usepackage{soul}
\usepackage{amssymb}

\definecolor{hlgreen}{HTML}{C0F4D6}
\definecolor{hlblue}{HTML}{B5D9F6}
\definecolor{hllightgray}{HTML}{F8F9FA}
\definecolor{hlgray}{HTML}{E6E6E6}
\definecolor{hlcaseblue}{HTML}{00FFFF}
\definecolor{hlcasegreen}{HTML}{00FF00}
\definecolor{hlcaseyellow}{HTML}{FFFF00}
\definecolor{darkgreen}{HTML}{006400}
\definecolor{lightred}{RGB}{255, 230, 230}
\newcommand{\hlc}[2][hlgreen]{\begingroup\sethlcolor{#1}\hl{#2}\endgroup}
\usepackage[table]{xcolor}

\author{
    Kerui Chen\textsuperscript{\rm 1}\thiswork,
    Jinglu Wang\textsuperscript{\rm 2},
    Xiaoyi Zhang\textsuperscript{\rm 2},
    Yan Lu\textsuperscript{\rm 2}\corresponding
}
\affiliations{
    \textsuperscript{\rm 1}Zhejiang University
    \textsuperscript{\rm 2}Microsoft Research Asia\\
}

\title{Beyond the Single Camera: Agentic Multi-View Reasoning \\ in Sports Video Understanding}

\newcommand{\qanumber}{3015}
\newcommand{\videonumber}{1022}
\newcommand{\ouragent}{SportMV-Agent}
\newcommand{\ourbench}{SportMV-Bench}

\begin{document}

\maketitle

\begin{abstract}
Recent Multimodal Large Language Models (MLLMs) achieve strong performance on single-view video understanding benchmarks. However, sports videos involve dense occlusion, rapid motion, and complex interactions that are difficult to resolve from a single viewpoint. In practice, sports events are recorded from multiple camera angles, providing complementary evidence used by referees. Yet, no existing benchmark evaluates MLLMs on multi-view sports video understanding.
To address this gap, we introduce \textbf{\ourbench}, a comprehensive benchmark built from official match recordings, through a dedicated pipeline combining LLM-based generation, MLLM-based verification, and human filtering to ensure quality and consistency. \ourbench~contains \videonumber~multi-view video bundles and \qanumber~question-answer pairs spanning 10 sports across three categories: Perception-Aware Recognition (PAR), Rule Event aware Interpretation (REI), and Adjudicative Decision Reasoning (ADR).
Our analysis shows that current MLLMs fail to effectively exploit multi-view information, with the bottlenecks lying in fine-grained visual perception and view selection rather than logical reasoning or domain knowledge.
We propose \textbf{\ouragent}, an agentic framework that orchestrates an iterative loop of active view selection, perception tool execution, and evidence-grounded reasoning, achieving a significant 15.61\% relative improvement over the strongest MLLM baseline.
\end{abstract}
\section{Introduction}
\label{sec:intro}

\begin{figure}[t]
\centering
\includegraphics[width=1.0\linewidth]{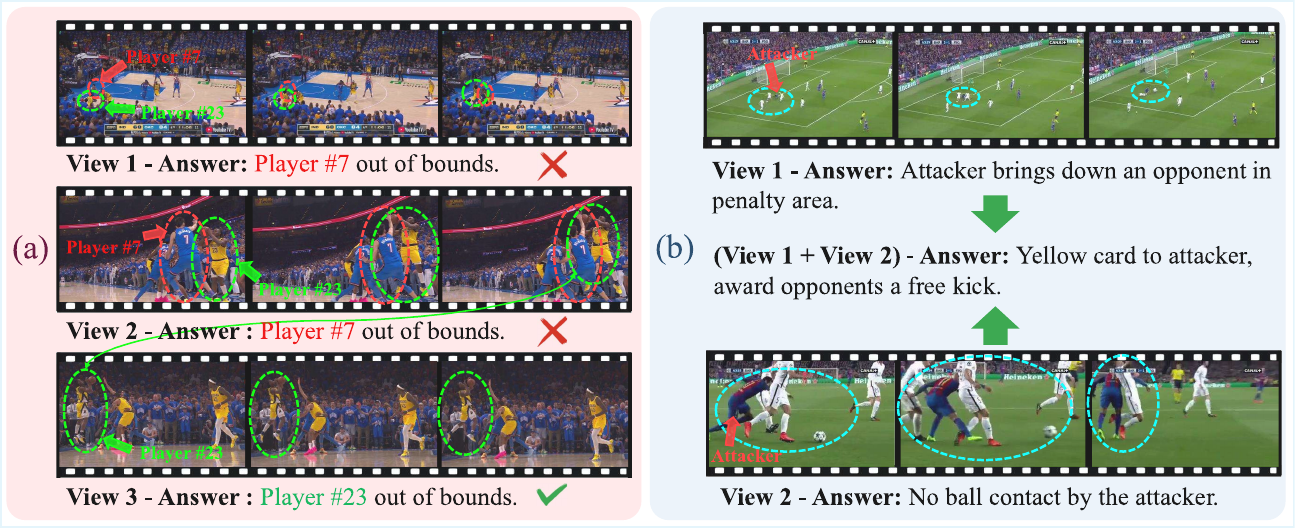}
\vspace{-5mm}
\caption{\textbf{Importance of multi-view perspectives in sports understanding.} (a) Single views (\texttt{View 1 \& 2}) can yield incorrect judgments (Player \#7 out) due to occlusion, while a more informative view (\texttt{View 3}) reveals the correct result (Player \#23 out). (b) Accurate decisions require aggregating complementary evidence from multiple views to resolve visual ambiguities: the attacker fouls an opponent (\texttt{View 1}) without touching the ball (\texttt{View 2}), awarding the defensive team a free kick.
}
\label{fig:intro}
\vspace{-2mm}
\end{figure}

Sports video understanding has received significant attention due to its wide applications in analytics~\cite{mackenzie2013performance}, coaching~\cite{groom2013application}, officiating support~\cite{X-VARS, held2024towards}, content retrieval~\cite{yu2003content}, and audience engagement~\cite{midoglu2024ai}.
Recent Multimodal Large Language Models (MLLMs)~\cite{bai2025qwen2,team2024gemini,a2025_gpt5,li2024llava, chen2026scaling} demonstrate strong capabilities in video understanding and reasoning, making them promising for sports video analysis.
However, sports videos pose unique challenges for MLLMs, such as 
rapid movements, dense player interactions, domain-specific rules, and the need for fine-grained subtle visual cues.
Several benchmarks have been developed to study sports video understanding, such as SoccerNet~\cite{giancola2018soccernet}, Sports-QA~\cite{Sports-QA}, and multi-sport datasets~\cite{SPORTR,SPORTU}. While these benchmarks advance holistic understanding of broadcast videos, they typically assume that a single camera view provides sufficient evidence for question answering.

In real-world, sports events are captured from multiple viewpoints to provide a complete perspective for referees and audiences.
Occlusion is common in sports: players often cluster around decisive moments, making critical actions hard to observe from a single view.
Moreover, athletes may intentionally obscure events through exaggerated contact or simulated falls, making single-view evidence not only incomplete but sometimes misleading.
As illustrated in Fig.~\ref{fig:intro}(a), using only \texttt{View 1} or \texttt{View 2} leads to an incorrect out-of-bounds decision because the decisive touch is occluded, while incorporating \texttt{View 3} reveals the correct outcome.
Beyond occlusion, some queries require combining complementary evidence across views that may even appear conflicting.
For example, in Fig.~\ref{fig:intro}(b), \texttt{View 1} localizes contact inside the penalty area, while \texttt{View 2} shows that the defender did not touch the ball. Only by combining both views can the correct decision be made: a yellow card to attacker, and award opponents a defensive free kick rather than a penalty.
These examples highlight the importance of multi-view reasoning for sports video understanding.
Professional officiating systems such as VAR, Instant Replay, and Hawk-Eye already rely on multi-camera setups for this reason.
However, existing sports QA benchmarks do not evaluate the ability of MLLMs to reason across multiple views of the same event, as they provide only single-view inputs.

To fill this gap, we introduce \textbf{\ourbench}, the first benchmark for multi-view sports video understanding. It spans 10 different sports and comprises \videonumber~multi-view video bundles (each with 2-5 diverse and asynchronous views) and \qanumber~QA pairs across three reasoning levels: Perception-Aware Recognition (PAR), Rule Event aware Interpretation (REI), and Adjudicative Decision Reasoning (ADR).
All videos are sourced from official public match recordings to ensure real-world authenticity. 
QA pairs are constructed through a four-stage pipeline combining automatic generation, multimodal verification and human verification to ensure correctness, consistency with the videos, and visual answerability.

Through extensive experiments and analysis, we find that current MLLMs perform poorly and fail to exploit multi-view information effectively. Providing all views jointly improves only 0.9 points over a random single view, whereas anchoring an oracle view yields a {13.4\%} relative gain, indicating that decisive evidence exists across views but current models fail to extract it. Moreover, providing annotated perception labels improves accuracy by {+21.72} points, highlighting the critical role of fine-grained perception in overall performance.
We also observe that chain-of-thought reasoning \emph{degrades} performance, and textual domain-specific knowledge provides little improvement. These findings suggest that the primary bottleneck lies not in logical reasoning or domain knowledge, but in visual perception.

Motivated by these findings, we propose \textbf{\ouragent}, an agentic multi-view reasoning framework.
An orchestrator coordinates an iterative loop of evidence collection and evidence-grounded reasoning: it actively selects informative views, invokes perception tools to extract fine-grained evidence, and accumulates findings until sufficient confidence is reached.
Compared to the baseline (GPT-4.1),  \ouragent~improves the overall performance from 59.12\% to \textbf{68.35\%}, a relative gain of \textbf{15.61\%}.

To sum up, our contributions are as follows:
\begin{itemize}
\item We present \textbf{\ourbench}, the first multi-view sports video understanding benchmark, consisting of \videonumber~multi-view video bundles and \qanumber~QA pairs across three reasoning levels (PAR, REI, ADR).
\item We conduct a comprehensive empirical study revealing that current MLLMs fail to exploit multi-view evidence, with the bottleneck in visual perception and view selection rather than logical reasoning or domain knowledge.
\item We propose \textbf{\ouragent}, an agentic framework whose orchestrator coordinates active view selection, perception tool execution, and evidence-grounded reasoning, with design directly motivated by the empirical findings.
\end{itemize}

\section{Related Work}
\label{sec:related}

\noindent\textbf{Sports understanding benchmarks.} 
Early sports understanding datasets primarily focused on action classification or action quality assessment in specific sports like soccer~\cite{giancola2018soccernet}, basketball~\cite{finesports}, diving~\cite{diving48}, gymnastics~\cite{finegym} and figure skating~\cite{FSD-10}.
Later datasets expanded the task from simple action classification to temporal action parsing~\cite{TAPOS,mcfs} and spatio-temporal localization~\cite{finesports}, and also introduced a wider variety of sports categories~\cite{multisports}.

With the advent of MLLMs~\cite{liu2024mmbench, fang2024mmbench, fu2025video, fu2025mme, yue2024mmmu}, question-answering (QA) benchmarks emerged as the standard paradigm for evaluating multimodal understanding in sports scenarios.
Single-sport datasets include SoccerNet-X Foul~\cite{held2023vars} and SoccerBench~\cite{rao2025multi} for recognition in soccer, FSBench~\cite{fsbench} for scoring in figure skating, and FineBadminton~\cite{finebadminton} for rally-level understanding in badminton.
Other datasets like Sports-QA~\cite{Sports-QA}, SPORTU~\cite{SPORTU} and SPORTR~\cite{SPORTR} extend evaluation to multiple sports, incorporating higher-level reasoning over rules and tactics.
However, these benchmarks typically offer a single-view video input, which cannot simulate or evaluate realistic scenarios where occlusions in one viewpoint necessitate joint reasoning over multiple views.

\noindent\textbf{Sports understanding methods.} 
Early approaches cast sports understanding as a classification or regression task, focusing on action recognition~\cite{finegym, multisports}, foul classification~\cite{X-VARS}, and action quality assessment~\cite{fsbench, diving48, parmar2019action}.
With the development of MLLMs, recent state-of-the-art models such as Qwen families~\cite{bai2025qwen2, wang2024qwen2}, GPT families~\cite{jaech2024openai, hurst2024gpt}, and Gemini families~\cite{team2024gemini, team2023gemini} show remarkable general capabilities in sports videos.
Many existing efforts treat them as a unified tool for tactical analysis~\cite{caron2023tacticalgpt, zhang2025chatmatch} and rule-based reasoning~\cite{held2024x}.
However, these approaches typically operate on a single viewpoint and assume that all information required to answer the question is fully contained within that view.
This assumption simplifies the problem compared to real-world sports scenarios, where key evidence is often distributed across multiple views and individual camera angles may suffer from occlusion.
As a result, the capabilities of MLLMs in realistic multi-view settings remain largely untested and underexplored, which motivates our evaluation of MLLMs on multi-view sports video understanding.

\begin{figure}[t]
\centering
\includegraphics[width=1.0\linewidth]{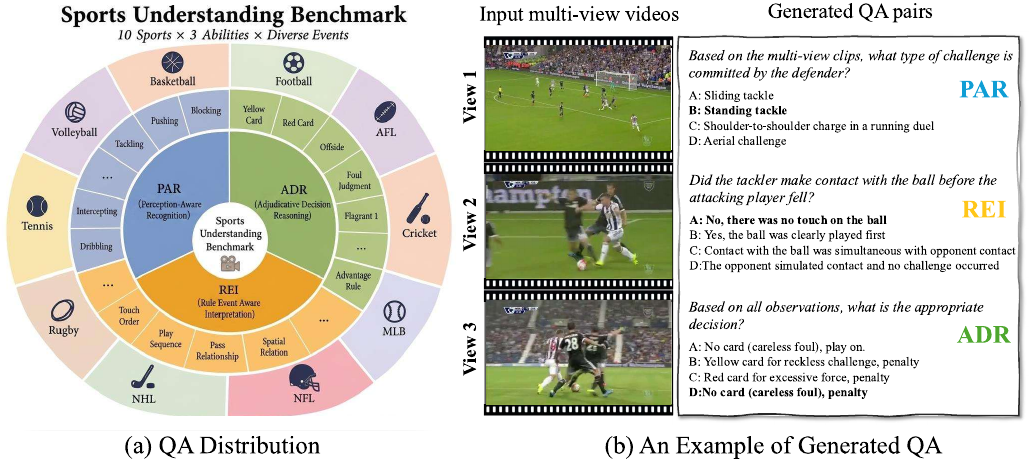}
\vspace{-6mm}
\caption{\textbf{Overview of \ourbench.} (a) The benchmark comprises ten sports categories and three types of questions, each designed to evaluate a different aspect of sports understading. (b) A real-match multi-view video example with its corresponding three types.}
\label{fig:static}
\vspace{-3mm}
\end{figure}
\section{Sport-MV Benchmark}
\label{sec:dataset}

In this section, we present \textbf{\ourbench}, a benchmark for multi-view sports video understanding designed to evaluate the capabilities of MLLMs in multi-view scenarios.
We detail its problem formulation, data construction pipeline and resulting data statistics.

\paragraph{Problem formulation.}
We consider the task of multi-view sports video understanding.
Let $\mathcal{V}=\{V_1, V_2, \ldots, V_N\}$ denote $N$ videos capturing the same sports event from different camera viewpoints, which may not be temporally synchronized. Given a natural-language question $q$, the goal is to infer the correct answer $a$ by aggregating evidence across the multi-view videos.

\begin{figure*}[t]
\centering
\includegraphics[width=0.9\linewidth]{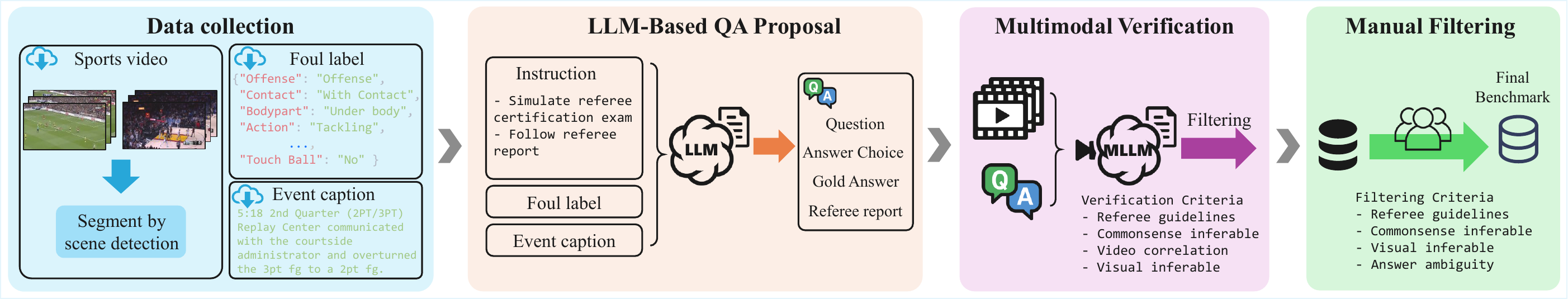}
\vspace{-2mm}
\caption{\textbf{Benchmark construction pipeline.} 
We construct the benchmark through four stages.
First, we collect paired videos and official referee reports from public video platforms.
Second, we employ a LLM to propose candidate QA pairs based on event captions and foul annotations, which helps mitigate the hallucination issue in MLLMs arising from redundant video content.
Third, we apply a MLLM to filter these pairs according to criteria such as video–QA consistency and commonsense verification.
Finally, human annotators further refine the dataset by removing ambiguous or unreasonable QA pairs.
}
\vspace{-3mm}
\label{fig:workflow}
\end{figure*}

\subsection{Dataset Construction}
To construct a reliable and semantically rich QA corpus, we design a pipeline leveraging LLMs and MLLMs, as illustrated in Figure~\ref{fig:workflow}, which comprises four stages: data collection, LLM-Based QA generation, multimodal verification and authoritative human annotations.

\noindent\textbf{Stage 1: Data collection.} 
We collect video sources and their corresponding official refereeing decisions from public video platforms.
To ensure sport diversity, \ourbench~spans ten sports: basketball, soccer, cricket, rugby, American football, ice hockey, baseball, tennis, volleyball and Australian football.
These sports jointly feature both global and local viewpoints, exhibit significant diversity across camera angles and scenes, and involve frequent occlusions with complex player interactions.
Across all sources, we deliberately prioritize contested and challenge plays to maximize difficulty.
To obtain multi-view observations of the same event, we employ a three-stage pipeline comprising viewpoint-change detection with \textit{PySceneDetect}, content and quality filtering using a vision-language model, and viewpoint clustering based on DINOv2~\cite{oquab2024dinov2} features.
After stage~1, for each data instance, we obtain a multi-view video set $\mathcal{V}$ and its associated referee report $R$.
Notably, these views are not temporally synchronized, further increasing difficulty by requiring cross-view alignment.
More details are provided in the supplementary material.

\noindent\textbf{Stage 2: LLM-Based QA proposal.}
Conditioning an MLLM directly on the multi-view clips for QA synthesis tends to produce hallucinated or redundant questions anchored to salient but non-decisive content.
We therefore ground QA generation in the referee report, an authoritative event-level answer source.
Specifically, we prompt an LLM to act as a referee certification exam setter.
For each report, the LLM proposes multiple-choice questions strictly grounded in the report text, with a gold answer extracted verbatim.
The stem is anonymized (no team, player, or date names) to prevent the answer from leaking through surface cues.
Distractors must share the same semantic type as the gold answer and be minimally contrasting variants drawn from non-deciding details, preventing shortcuts based on type mismatch.
The prompt forbids external knowledge and hallucinated entities, and enforces a fixed JSON output containing the question, options, and gold answer, denoted as the tuple $(q, \mathcal{A}, a^*)$.
Each synthesized item is linked to its corresponding multi-view clips for later  multimodal verification.

\noindent\textbf{Stage 3: Multimodal verification.}
Despite the constraints in Stage~2, some synthesized items still deviate from the referee report, are answerable by commonsense alone, are unrelated to the video, or lack sufficient visual evidence.
We therefore employ a strong MLLM (GPT-5~\cite{a2025_gpt5}) as an auditor.
For each candidate tuple $(q, \mathcal{A}, a^*)$ with the referee report $R$ and multi-view clips $\mathcal{V}$, the auditor 1) verifies whether the gold answer is consistent with $R$; 2) determines whether the item is answerable by commonsense alone, 3) checks whether the QA pair is semantically aligned with the video, and 4) assesses whether the answer can be inferred from the video. We define a quality score $\in [0,10]$ to quantify this verification process.
Items that fail the consistency or commonsense check, or fall below the confidence threshold, are filtered out.
To reduce artifacts, the audit is repeated three times with the view order shuffled, and an item must pass all three runs to be accepted.

\noindent\textbf{Stage 4: Authoritative manual filtering.}
Finally, we recruit five expert annotators to review the remaining items to filter out ambiguous, non-visual, weakly grounded, or ill-posed questions, and to refine the wording and options so that the gold answer is clear and unique.
The resulting corpus achieves high linguistic and multimodal alignment, providing a robust basis for evaluating multi-view, fine-grained visual–textual understanding in sports officiating.

\subsection{Dataset Statistics}
Our final \ourbench~consists of \textbf{\qanumber} QA pairs, each pairing a natural-language question with multi-view clips of the same event and requiring cross-view evidence aggregation and reasoning to answer.
As shown in Fig.~\ref{fig:static}, we categorize questions into three types, PAR, REI and ADR, which account for \textbf{35.23\%}, \textbf{36.98\%}, and \textbf{27.79\%} of the corpus, respectively.
These categories reflect increasing levels of reasoning and dependence on multi-view evidence.
We detail each category as follows:
\begin{itemize}
  \item \textbf{Perception-Aware Recognition (PAR).} Evaluates low-level perceptual grounding of \emph{what happened} across views (e.g., tackle, block, push).
  The focus is on robust action identification under occlusion, motion blur, or adverse framing, benefiting from cross-view corroboration. 

  \item \textbf{Rule Event Aware Interpretation (REI).} Probes mid-level, event/rule-sensitive reasoning: the question hinges on a rule or event-based anchors specified in the query.
  The model must localize these anchors in space and time across views and then obtain rule or event-related answers about what happens before or after these anchors.

  \item \textbf{Adjudicative Decision Reasoning (ADR).} Assesses high-level normative inference by synthesizing multi-camera views to conclude \emph{what should be called}.
  This involves localizing the foul context (e.g., inside vs.\ outside the box), estimating severity and intent, and mapping the situation to the appropriate sanction (yellow/red card, free kick, penalty, flagrant) without any explicit hints, thereby rigorously testing cross-view spatiotemporal integration and rule-grounded decision making.
\end{itemize}

\noindent In summary, \ourbench~is the first benchmark that requires MLLMs to jointly reason across multiple asynchronous camera views of the same sports event.
Its three-level question hierarchy (PAR, REI, ADR) mirrors the cognitive process of real-world sports officiating, from perceiving actions, to interpreting events under rules, to rendering adjudicative decisions.

\section{Benchmark Analysis and Findings}
\label{sec:analysis}
In this section, we establish a baseline evaluation protocol for MLLMs on \ourbench~and conduct a systematic analysis to diagnose performance bottlenecks.

\subsection{Baseline Evaluation Protocol}
Given the multi-view video set $\mathcal{V}$ and question $q$, an MLLM predicts the answer $\hat{a}$ by aggregating evidence across views.
A straightforward baseline treats the multi-view bundle as a single concatenated input, delegating all view selection and cross-view fusion to the MLLM.
Specifically, we pack all views with view-identifier tokens and question $q$ into a unified sequence $S$ and feed it to the MLLM:
\begin{equation}
    \label{eq:baseline}
    \hat{a} = \texttt{MLLM}(S), \\
    \\
    S = [\, q ;\; \langle\mathrm{V}\,1\rangle\, X_1 ;\; \dots ;\; \langle\mathrm{V}\,N\rangle\, X_N \,],
\end{equation}
where $X_i$ denotes uniformly sampled frames from view $V_i$.
For multiple-choice QA, the prediction follows likelihood maximization:
$\hat{a} = \arg\max_{a_k \in \mathcal{A}} \log p_{\theta}(a_k \mid S)$.
This baseline imposes no explicit structure for cross-view reasoning; the MLLM must implicitly identify informative viewpoints, discount misleading perspectives, and integrate complementary cues within its fixed context budget.

We evaluate a broad set of state-of-the-art open-source and proprietary MLLMs under this protocol. The detailed experimental setup and full quantitative results are provided in experiment section. Overall, even the strongest models perform poorly, revealing substantial challenges in this task.


\begin{table}[t]
\centering
\footnotesize
\setlength{\tabcolsep}{6pt}
\begin{adjustbox}{width=0.9\linewidth}
\begin{tabular}{c|cccccc}
\toprule
{Setting} & {PAR} & {REI} & {ADR} & {Overall} \\
\midrule
1-View (Random) & 41.73 & 52.58 & 27.84 & 42.35 \\
1-View (Best) & 44.41 & 56.27 & 32.46 & 45.91 \\
M-View (Guided) & 48.36 & 59.86 & 33.21 & 49.04 \\
\rowcolor{hlblue}
M-View (All) & 42.63 & 52.41 & 29.97 & 43.26\\
\midrule
+ Knowledge & 43.59 & 56.68 & 31.82 & 45.43 \\
+ Oracle Tool & 88.71 & 61.09 & 38.15 & 64.98 \\
+ COT & 39.82 & 44.53 & 29.18 & 39.32 \\
\bottomrule
\end{tabular}
\end{adjustbox}
\vspace{-2mm}
\caption{Analysis of view configurations and different strategies on MLLM performance. The \hlc[hlblue]{highlighted} row is Qwen3-VL-30B under our baseline evaluation protocol, and the bottom three rows build on it with different strategies.}
\vspace{-4mm}
\label{finding}
\end{table}

\begin{figure*}[t]
\centering
\includegraphics[width=0.9\linewidth]{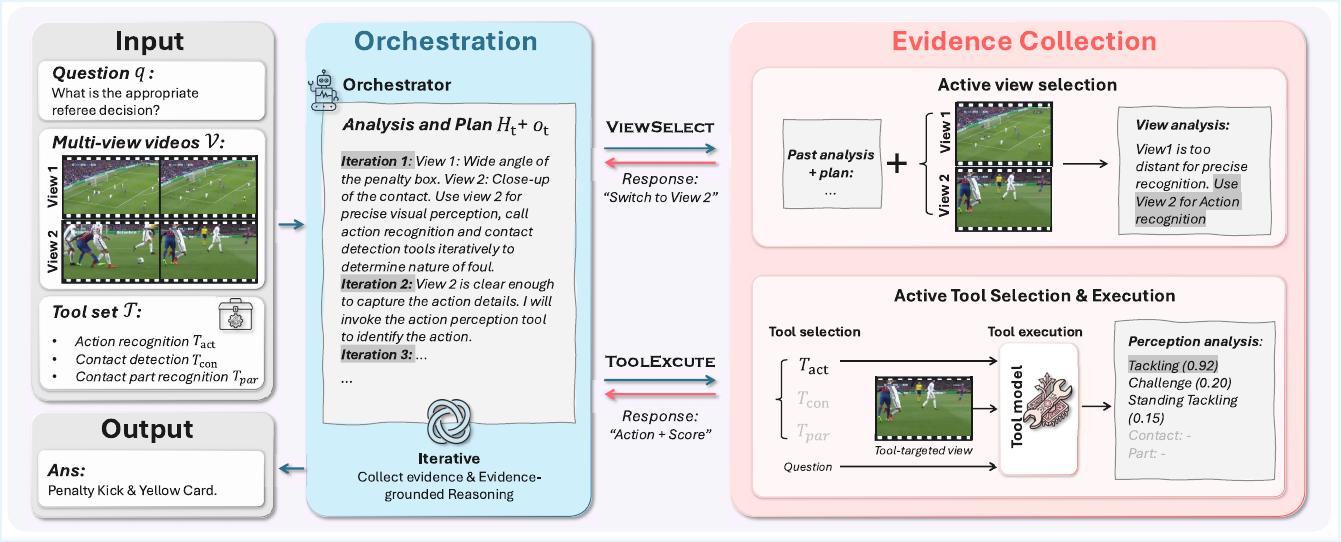}
\vspace{-2mm}
\caption{\textbf{Overview of \ouragent.} The orchestrator coordinates an iterative loop: at each step it selects views, invokes perception tools, analyzes evidence reliability, and decides whether to continue or emit the final answer.}
\label{fig:method}
\vspace{-3.5mm}
\end{figure*}

\subsection{Bottleneck Analysis}
Beyond general video understanding, multi-view sports understanding introduces two additional challenges: (1) effective exploitation of multi-view information, and (2) robust sports-specific reasoning.
We analyze these aspects through controlled experiments and extract insights that guide our method design.

\subsubsection{Can MLLMs effectively leverage multi-view information?}
To examine how multi-view information affects MLLM performance, we evaluate MLLMs under four view configurations that vary in two dimensions: view quantity (single vs.\ multiple) and view quality (random vs.\ best vs.\ guided).
\begin{itemize}
    \item \textbf{1-View (Random)}: a single randomly sampled viewpoint;
    \item \textbf{1-View (Best)}: the most suitable viewpoint for the question, annotated by a strong MLLM (GPT-5);
    \item \textbf{M-View (All)}: all available viewpoints provided jointly;
    \item \textbf{M-View (Guided)}: multiple views provided, with explicit instruction to prioritize the best view while using others as auxiliary references.
\end{itemize}
All configurations follow the baseline evaluation protocol in Eq.~\ref{eq:baseline}, differing only in the active views included in $S$ and in prompting.
We evaluate Qwen3-VL-30B, with results shown in Tab.~\ref{finding}, and observe two key findings:

\noindent \textbf{1)} \textit{Quantity of views $\neq$ quality of information.}
\textbf{M-View (All)} yields only marginal gains over \textbf{1-View (Random)} and remains substantially worse than \textbf{1-View (Best)}.
Redundant, occluded, or misleading angles introduce visual noise that offsets the benefit of complementary perspectives.

\noindent \textbf{2)} \textit{Active view selection is critical.}
\textbf{M-View (Guided)} achieves the best performance. The optimal strategy is not to ingest more views, but to selectively focus on and aggregate the informative ones.
Current MLLMs lack explicit mechanisms for autonomous view selection, structured cross-view aggregation, and evidence-aware reasoning.

\subsubsection{What can improve sports-specific reasoning?}
We further examine representative strategies targeting reasoning bottlenecks:(1) \textit{Chain-of-Thought} (\texttt{COT}), a widely adopted strategy for enhancing reasoning in LLMs;
(2) \textit{Sports-specific knowledge} (\texttt{SP-Knowledge}), which provides domain knowledge that the MLLM may lack;
and (3) \textit{Fine-grained visual perception} (\texttt{FG-Perception}), which supplies fine-grained action and contact recognition and grounding information.
The effects of these strategies are shown in Tab.~\ref{finding}.
We draw the following three lessons:

\noindent \textbf{1)} \textbf{\textit{COT solely cannot improve performance.}}
Despite its proven effectiveness in many language and vision-language tasks, COT reasoning surprisingly leads to a decline in accuracy for Qwen3-VL-30B on our benchmark.
This suggests that in complex multi-view sports scenes, sequential reasoning without reliable visual grounding amplifies hallucination rather than improving decision quality.

\noindent \textbf{2)} \textbf{\textit{Domain knowledge is not the primary bottleneck.}}
We provide detailed sports-specific guidelines for each option, generated by GPT-5 to highlight decisive cues.
As shown in Tab.~\ref{finding}, performance gains are negligible, indicating that conceptual knowledge is not the limiting factor.

\noindent \textbf{3)} \textbf{\textit{Fine-grained visual perception is decisive.}}
When supplied with annotated action and contact, performance improves dramatically. This indicates that precise perception of player actions and interactions constitutes a primary bottleneck for MLLMs in multi-view sports understanding. High-level reasoning strategies alone cannot compensate for deficiencies in low-level perceptual accuracy.

\noindent\textbf{Summary.}
Our analysis reveals three core bottlenecks for baseline MLLMs:
(1) the inability to autonomously select informative views, 
(2) fragile reasoning performance that is highly sensitive to perceptual errors, and
(3) insufficient fine-grained visual perception, particularly for action recognition and contact detection.

\section{\ouragent}
\label{sec:method}
As analyzed in the above section, existing MLLMs struggle with multi-view sports understanding.
A single forward pass forces the model to passively process all perspectives at once, including occlusion and misleading views.
We instead adopt an agentic approach: \textbf{\ouragent}~actively decides which view to examine, invokes specialized perception tools to extract concrete evidence, and iteratively accumulates findings until sufficient confidence is reached.

\begin{figure}[t]
\centering
\vspace{-4mm}
\begin{algorithm}[H]
\small
\caption{SportMV-Agent workflow.}
\label{alg:agent}
\begin{algorithmic}[1]
\REQUIRE $q$, $\mathcal{V}\!=\!\{V_1,\ldots,V_N\}$, $\mathcal{T}$, $\pi_\theta$
\ENSURE Predicted answer $\hat{a}$
\STATE $\mathcal{V}^{\mathrm{act}}_0\!\leftarrow\!\mathcal{V}$;\; $E_0\!\leftarrow\!\emptyset$;\; $t\!\leftarrow\!1$
\hfill{\scriptsize$\triangleright$ \textit{initialize}}
\WHILE{$t \leq t_{\max}$}
    \STATE $s_t \!\leftarrow\! (q, \mathcal{V}^{\mathrm{act}}_{t-1}, \mathcal{E}_{t-1})$
    \hfill{\scriptsize$\triangleright$ \textit{construct state}}
    \STATE $(H_t, o_t) \!\leftarrow\! \pi_\theta(s_t)$ \quad{\scriptsize$\triangleright$ \textit{generate thought + op.}}
    \IF{sufficient confidence from $H_t$}
        \RETURN $\hat{a} \leftarrow o_t$
        \hfill{\scriptsize$\triangleright$ \textit{emit final answer}}
    \ELSIF{$o_t$ is \textsc{ViewSelect}}
        \STATE $\mathcal{V}^{\mathrm{act}}_{t} \!\leftarrow\! o_t(\mathcal{V})$
        \hfill{\scriptsize$\triangleright$ \textit{select views}}
        \STATE $E_t \!\leftarrow\! \pi_\theta(\mathcal{V}^{\mathrm{act}}_{t})$
        \hfill{\scriptsize$\triangleright$ \textit{collect evidence}}
    \ELSIF{$o_t$ is \textsc{ToolExec}}
        \STATE ($T^{\mathrm{act}}\!,V^{\mathrm{tool}}) \!\leftarrow\! o_t(\mathcal{T},\mathcal{V})$
        \hfill{\scriptsize$\triangleright$ \textit{select tool, target view}}
        \STATE $E_t \!\leftarrow\! \pi_\theta(T^{\mathrm{act}}(V^{\mathrm{tool}}))$
        \hfill{\scriptsize$\triangleright$ \textit{collect evidence}}
    \ENDIF
    \STATE $\mathcal{E}_{t} \!\leftarrow\! \mathcal{E}_{t-1} \!\cup\! \{E_t\}$;\; $t \!\leftarrow\! t\!+\!1$
    \hfill{\scriptsize$\triangleright$ \textit{accumulate evidence}}
\ENDWHILE
\end{algorithmic}
\end{algorithm}
\vspace{-6mm}
\end{figure}

\noindent\textbf{Overview.}
As illustrated in Fig.~\ref{fig:method}, \ouragent~takes as input a question $q$, multi-view videos $\mathcal{V}$, and a set of perception tools $\mathcal{T}$, and outputs the predicted answer $\hat{a}$.
The framework consists of an orchestrator that coordinates an iterative loop of \emph{evidence collection} and \emph{evidence-grounded reasoning}, continuing until sufficient evidence is gathered for a confident prediction.

\subsection{Orchestration}
\label{sec:orchestrator}
The orchestrator $\pi_\theta$ serves as the central controller of \ouragent. The full procedure is summarized in Alg.~\ref{alg:agent}.
It maintains a reasoning state $s_t = (q, \mathcal{V}^{\mathrm{act}}_{t-1}, \mathcal{E}_{t-1})$ comprising the question, the currently active views, and the evidence accumulated from previous steps.
At each iteration, given $s_t$, the orchestrator generates a thought $H_t$ and an operation $o_t$. $H_t$ represents a textual analysis and plan for the next step. $o_t$ is either the final answer or an evidence collection operation for next step, determined by the confidence of $H_t$. If the confidence is insufficient, the orchestrator triggers the evidence-collection operation $o_t$ to obtain new evidence $E_t$; otherwise, it outputs the final answer $\hat{a}$.

\subsection{Evidence Collection}
\label{sec:evidence_collection}
Evidence is collected through two complementary operations, chosen by the orchestrator according to the current state $s_t$: \emph{active view selection}, which decides \emph{where to look}, and \emph{active tool execution}, which decides \emph{what to extract}.

\noindent\textbf{Active view selection.}
In complex sporting events, the ``optimal'' view is dynamic and context-dependent.
When the orchestrator determines that the current view suffers from occlusion, motion blur, or lacks critical detail, it triggers a view-selection operation $o_t = \textsc{ViewSelect}$ to select informative angles $\mathcal{V}^{\mathrm{act}}_t \subseteq \mathcal{V}$.
The system starts with access to the full set of available views, denoted as $\mathcal{V}^{\mathrm{act}}_0 = \mathcal{V}$.
Whenever a view switch occurs, the video segment from the newly selected view is integrated into the agent's observation context.

\noindent\textbf{Active tool selection and execution.}
Fine-grained perception is demonstrated to be critical for multi-view sports understanding.
We equip the framework with a Perception Toolbox $\mathcal{T}$ that delivers sports-specific expertise as \emph{perception} rather than text. The orchestrator invokes these tools through the operation $o_t= \textsc{ToolExec}$, which selects the appropriate tool and target view based on the current state.
The toolbox is extendable and currently includes three fundamental tools designed for sports analysis:
\begin{itemize}
\item \textbf{Action Recognition} ($T_\texttt{act}$): outputs a ranked list of action names with confidence scores.
\item \textbf{Contact Detection} ($T_\texttt{con}$): outputs a binary judgment of whether physical contact occurs between two players.
\item \textbf{Contact Part Detection} ($T_\texttt{par}$): triggered when contact is detected, outputs the body region involved.
\end{itemize}
For example, given a football play question, the orchestrator may start from a global view to identify the foul area. Realizing that the specific action type and physical contact is obscured, it triggers a view switch to a close-up sideline angle, invokes $T_\texttt{act}$ to categorize the type of action, and then invokes $T_\texttt{con}$ to determine the situation of contact before emitting a final decision.
\section{Experiment}
\label{sec:experiment}

In this section, we evaluate both baseline MLLMs and our proposed \ouragent~on \ourbench. We first present the main results comparing \ouragent~against state-of-the-art MLLMs, followed by ablation studies that validate the contribution of each component.

\begin{table}[t]
\centering
\footnotesize
\setlength{\tabcolsep}{6pt}
\begin{adjustbox}{width=1.0\linewidth}
\begin{tabular}{c|cccccc}
\toprule
{Model} & {PAR} & {REI} & {ADR} & {Overall} \\
\midrule
LLaVa-Next~\cite{li2024llava} & 47.18 & 56.50 & 32.85 & 47.43\\ 
Qwen2.5-VL-72B~\cite{bai2025qwen2} & 45.31 & 53.99 & 35.75 & 45.34\\
Qwen3-VL-30B~\cite{qwen3technicalreport} & 42.63 & 52.41 & 29.97 & 43.26\\
InternVL2.5~\cite{chen2024expanding} & 45.16 & 56.73 & 34.42 & 45.79\\
GLM-4.5V~\cite{v2025glm} & 48.36 & 56.95 & 33.95 & 48.40\\
DeepSeek-VL~\cite{deepseek-vl} & 32.31 & 37.46 & 27.17 & 32.88\\
GPT-4o~\cite{gpt-4o} & 52.25 & 59.55 & 41.80 & 52.31\\
GPT-4.1~\cite{gpt-4o} & \underline{58.23} & \underline{65.99} & \underline{49.09} & \underline{59.12}\\
\midrule
\ouragent & \textbf{68.57} & \textbf{72.56} & \textbf{62.22} & \textbf{68.35}\\
\bottomrule
\end{tabular}
\end{adjustbox}
\vspace{-2mm}
\caption{Quantitative comparisons of the existing state-of-the-art MLLMs on the SportMV-Bench. The best and second best results are \textbf{bold} and \underline{underlined}. All the numbers are presented in \%.}
\label{main}
\vspace{-3mm}
\end{table}

\noindent\textbf{Baseline MLLMs.}
We evaluate eight state-of-the-art MLLMs on \ourbench~using the protocol in benchmark analysis, including open-source models (LLaVA-Next~\cite{li2024llava}, Qwen2.5-VL-72B~\cite{bai2025qwen2}, Qwen3-VL-30B~\cite{qwen3technicalreport}, InternVL2.5-38B~\cite{chen2024expanding}, GLM-4.5V~\cite{v2025glm}, DeepSeek-VL~\cite{deepseek-vl}) and proprietary models (GPT-4o~\cite{gpt-4o}, GPT-4.1).

\noindent\textbf{Implementation details.}
Videos are processed at 2 fps. We report exact-match accuracy averaged over three runs. Each model receives the same prompt and outputs the answer within \texttt{<answer>...</answer>} tags, parsed via regex (with gpt-4.1-mini as fallback). In \ouragent, GPT-4.1 serves as the orchestrator, while Qwen2.5-VL-72B implements the perception tools, returning top-3 predictions with confidence scores. Additional details are provided in the supplementary material.

\begin{figure*}[t]
\centering
\includegraphics[width=0.9\linewidth]{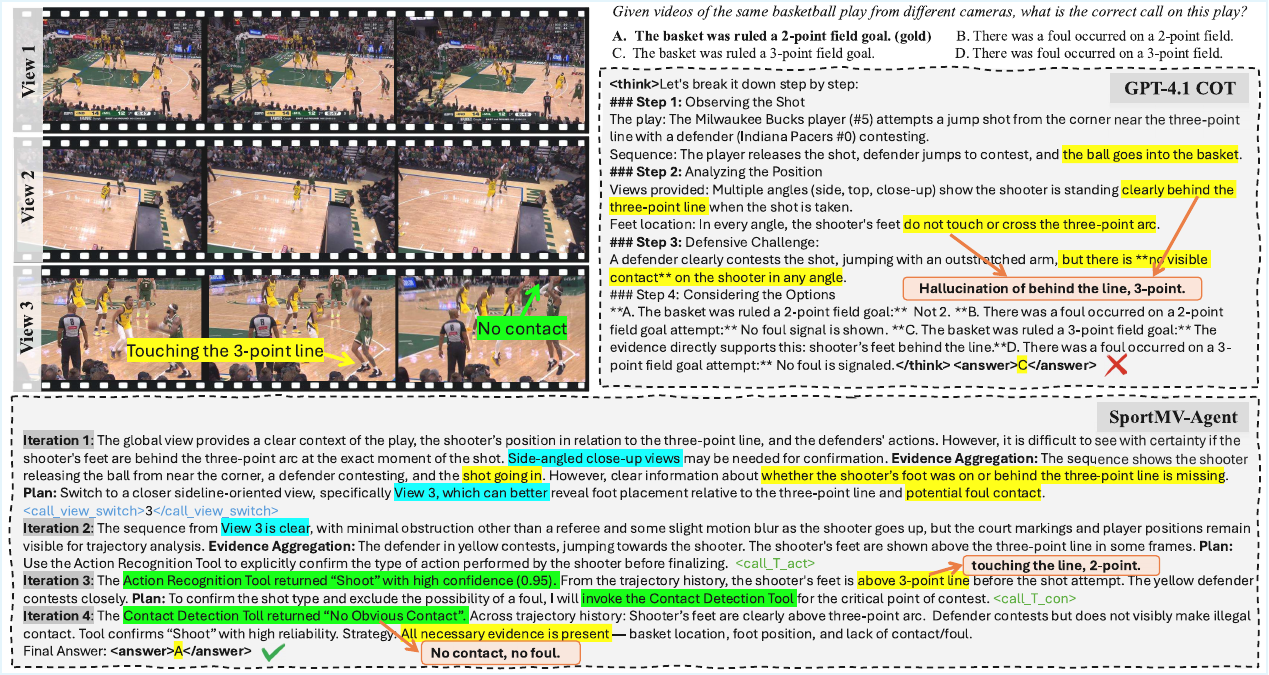}
\vspace{-3mm}
\caption{\textbf{Qualitative example.} Comparison between the reasoning outputs of \ouragent~and the COT baseline. The COT strategy suffers from visual error accumulation: due to occlusions in \textbf{View 1} and \textbf{View 2}, it fails to consider the critical evidence visible in \textbf{View 3}, incorrectly classifying the shot as a 3-pointer. In contrast, \ouragent~actively selects the most informative viewpoint (\textbf{View 3}), capturing the player's foot-on-line detail. By combining view selection with perception tools, the framework correctly identifies the event as a 2-point field goal without foul. Key clues are highlighted in \hlc[hlcaseyellow]{yellow}, view analysis in \hlc[hlcaseblue]{blue}, tool perception in \hlc[hlcasegreen]{green}.}
\label{fig:experiment}
\vspace{-3mm}
\end{figure*}

\subsection{Main Result}
\label{sec:main_result}
We report the main quantitative results in Tab.~\ref{main}.
Among all the baseline MLLMs, GPT-4.1 achieves the best performance at 59.12\%, while GLM-4.5V is the strongest open-source model (48.40\%), leaving a substantial gap between open- and closed-source models.
Notably, models with strong general video understanding capabilities, such as Qwen3-VL-30B, show a performance drop on our multi-view benchmark, highlighting the unique challenges of multi-perspective sports reasoning.
A category-wise results reveal that most models excel in REI but struggle with PAR and ADR, with average performance gaps of about 8.5\% and 19.3\%, respectively.
As discussed in benchmark analysis, REI often depends on a single decisive cue, whereas PAR is hindered by severe occlusions and ADR requires integrating evidence across multiple viewpoints together with rule-based reasoning.
\ouragent, using GPT-4.1 as the orchestrator, significantly boosts performance from 59.12\% to \textbf{68.35\%}, a relative improvement of 15.61\%.
This validates that the agentic design, effectively bridges the multi-view reasoning gap that standard MLLMs fail to address.

A qualitative comparison with the COT baseline is shown in Fig.~\ref{fig:experiment}. COT suffers from visual error accumulation, whereas \ouragent~actively selects the most informative view to capture critical details and reach the correct decision. This example demonstrates how view switching and tool invocation mitigates occlusion and improves multi-view evidence integration.

\subsection{Ablation Study}
\label{sec:ablation}
\begin{table}[t]
\centering
\setlength{\tabcolsep}{3pt}
\footnotesize
\newcommand{\up}[1]{{\scriptsize\color{red}\raisebox{-1pt}{+#1}}}
\newcommand{\dn}[1]{{\scriptsize\color{green!60!black}\raisebox{-1pt}{-#1}}}
\begin{adjustbox}{width=1.0\linewidth}
\begin{tabular}{ccccc|llll}
\toprule
{} & {Orchestrator} & Tool & Top-K & VS & {PAR} & {REI} & {ADR} & {Overall} \\
\midrule
1 & GPT-4.1 & - & - & - & 58.23\dn{10.34} & 65.99\dn{6.57} & 49.09\dn{13.13} & 59.12\dn{9.23} \\
2 & GPT-4o & Qwen & Top-3 & \checkmark & 57.44\dn{11.13} & 65.30\dn{7.26} & 46.01\dn{16.21} & 57.36\dn{10.99} \\
3 & GPT-4.1 & Qwen & Top-3 & $\times$ & 61.37\dn{7.20} & 64.54\dn{8.02} & 53.69\dn{8.53} & 60.63\dn{7.72} \\
4 & GPT-4.1 & GPT-4.1 & Top-3 & \checkmark & 71.43\up{2.86} & 71.95\dn{0.61} & 61.48\dn{0.74} & 68.76\up{0.41} \\
5 & GPT-4.1 & Qwen & Top-1 & \checkmark & 67.97\dn{0.60} & 70.17\dn{2.39} & 57.33\dn{4.89} & 66.18\dn{2.17} \\
\rowcolor{hlblue}
6 & GPT-4.1 & Qwen & Top-3 & \checkmark & 68.57 & 72.56 & 62.22 & 68.35 \\
\bottomrule
\end{tabular}
\end{adjustbox}
\vspace{-2mm}
\caption{Ablation study on the key elements in \ouragent. The VS column indicates whether the framework uses active view selection. {\color{red}Red}/{\color{green!60!black}green} numbers indicate improvement/decline relative to our implementation of \ouragent~(\hlc[hlblue]{row 6}).}
\vspace{-3mm}
\label{tab:ablation_new}
\end{table}

We conduct ablation studies to evaluate the contributions of key components in \ouragent, summarized in Tab.~\ref{tab:ablation_new}. The performance of \ouragent~is shown in Row 6.

\noindent\textbf{Impact of the orchestrator model.}
As shown in Row 2, replacing GPT-4.1 with the less capable GPT-4o as the orchestrator results in a performance drop of 10.99\%, underscoring the critical role of a strong reasoning backbone for effective evidence-grounded multi-view understanding.

\noindent\textbf{Impact of active view selection.}
As shown in Row 3, disabling the view switching module leads to a performance drop of 7.72\%. By actively switching to viewpoints with fewer occlusions, the orchestrator provides higher-quality visual inputs for perception tools, thereby reducing perceptual ambiguity and limiting error propagation. This observation is consistent with the view-selection bottleneck revealed in our benchmark analysis.

\noindent\textbf{Impact of perception tool models.}
As shown in Row 4, replacing the Qwen perception tool with GPT-4.1 yields only a marginal overall improvement of 0.41\%. Only the PAR category that focuses more on perception improves by 2.86\%, while the other two categories even decline slightly. This suggests that the framework is robust to the choice of perception models and can leverage more cost-effective alternatives without significant loss in performance.

\noindent\textbf{Robustness to tool response noise.}
As shown in Row 5, restricting the retrieved results from top-3 candidates to top-1 results leads to a performance drop of 2.17\%. This suggests that the framework possesses an inherent self-correction mechanism; by providing a broader candidate set, the orchestrator's evidence analysis can filter out noise and rectify potential retrieval errors.
\section{Conclusion}
\label{sec:conclusion}
In this paper, we introduce SportMV-Bench, the first multi-view benchmark designed to evaluate MLLMs on multi-view sports video understanding.
SportMV-Bench is built from real-world sports match videos, covers 10 different sports and three primary question types, and contains \qanumber QA pairs that have been dual filtered by an MLLM and human annotators.
Experiments reveal that current models often fail to select decisive viewpoints, perform precise perception, and aggregate complementary evidence, leading to large gaps in multi-view sports understanding.
To address these challenges, we present SportMV-Agent, an agentic framework equipped with an active view selection mechanism and the ability to invoke various perception tools.
We hope this challenging benchmark will foreground the importance of multi-view evidence in sports videos and promote robust multi-view reasoning for sports understanding in real-world applications. 
\newpage

\clearpage

\bibliography{aaai2027}

\clearpage
\setcounter{page}{1}
\setcounter{section}{0}
\setcounter{table}{0}
\setcounter{figure}{0}
\renewcommand\thesection{\Alph{section}}
{
\centering
\Large
\textbf{Beyond the Single Camera: Agentic Multi-View Reasoning in Sports Video Understanding}\\
\vspace{0.5em}Supplementary Material \\
\vspace{1.0em}
}

In this supplementary material, we present the following:
\begin{itemize}
    \item Discussion of MLLM-Based QA Generation

    \item More Qualitative Example

    \item More Implementation Details

    \item More Data Construction Details

    \item Dataset QA Examples

\end{itemize}

\section{Discussion of MLLM-Based QA Generation}
\begin{figure}[t]
\centering
\includegraphics[width=1.0\linewidth]{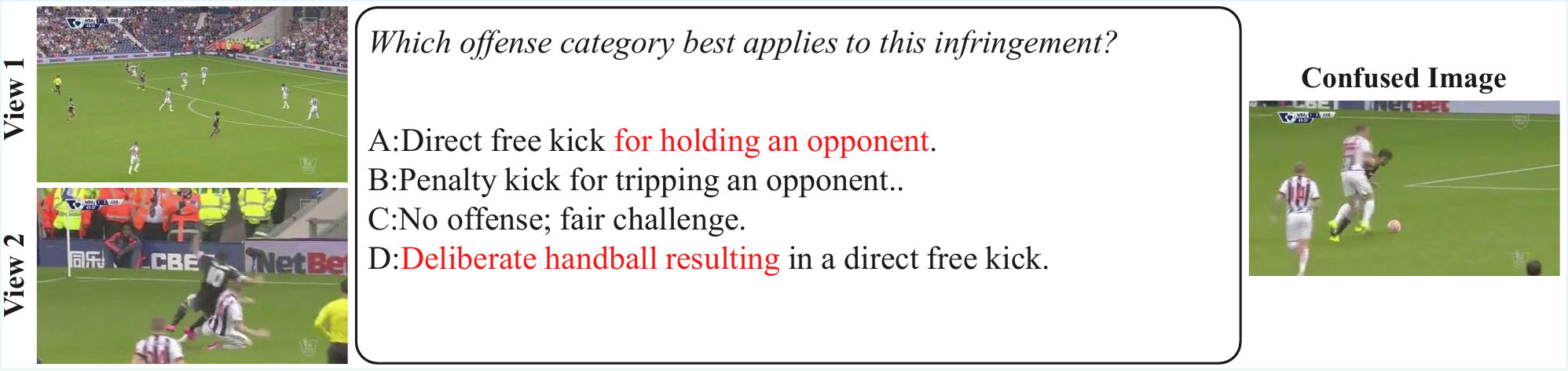}
\caption{Hallucinated Answer Choice Example. The direct free kick is actually caused by a sliding tackle, but the MLLM introduces wrong visual information into the options, so none of the options is correct.}
\label{supple:mllm_case1}
\end{figure}

\begin{figure}[t]
\centering
\includegraphics[width=1.0\linewidth]{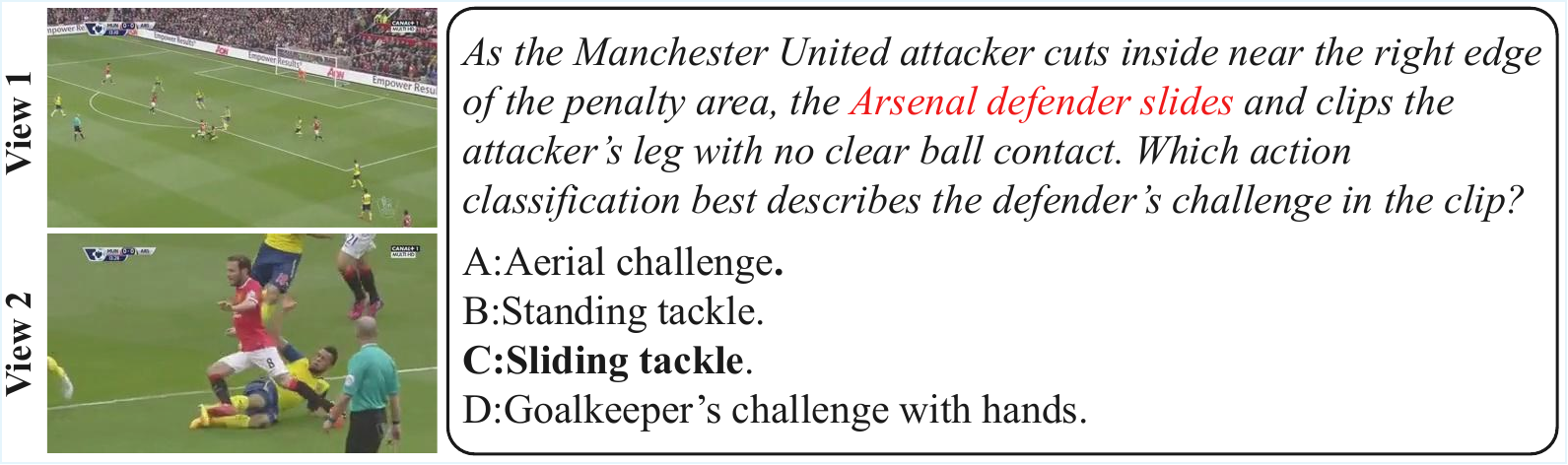}
\caption{Easily Answerable Item Example. The key cue “slide” already appears in the question, so the MLLM can easily produce the correct answer without real understanding, which is undesirable for evaluation.}
\label{supple:mllm_case2}
\end{figure}

\begin{figure}[t]
\centering
\includegraphics[width=1.0\linewidth]{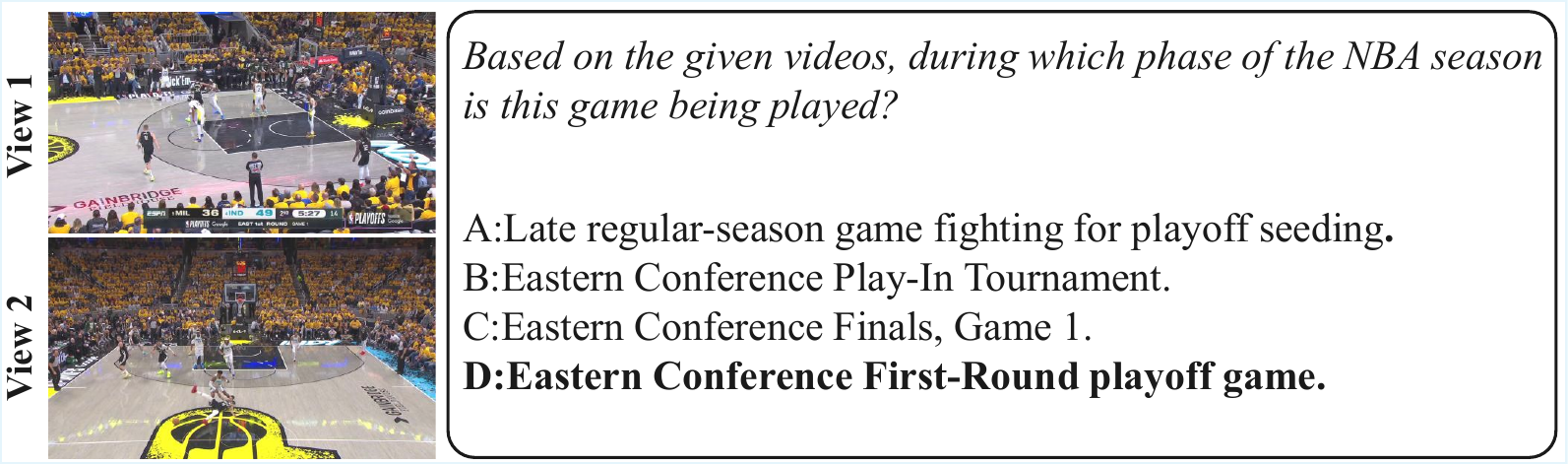}
\caption{General Video Question Example. The question and options are all correct, but is unrelated to the sports knowledge.}
\label{supple:mllm_case3}
\end{figure}

As mentioned in the main paper, using MLLMs as a QA generator can yield undesirable items once visual input is introduced.
Beyond the failure mode discussed in the main paper, we observe three additional types:
\begin{itemize}
\item \textbf{Hallucinated Answer Choice.} 
The MLLM sometimes misperceives the visual scene, and such erroneous visual information is mixed with the correct signals from the referee report, giving rise to spurious but plausible answer options.
As shown in Fig.~\ref{supple:mllm_case1}, the referee report clearly labels the decision as a free kick.
However, after incorporating visual information, the MLLM incorrectly introduces spurious events (e.g., a holding foul in the confused image) into the options, resulting in no candidate corresponding to the true decision.

\item \textbf{Easily Answerable Item.}
Some questions are supposed to require first perceiving the video and then reasoning over the context.
However, the MLLM may directly include the information that should be visually perceived in the question, making the resulting item overly simple.
As shown in Fig.~\ref{supple:mllm_case2}, although the correct answer is a sliding tackle, the MLLM explicitly encodes the observed visual cue (“defender slide”) in the question.
This causes the question to be trivial and easily answerable even without watching the video.
Meanwhile, when we use the same MLLM (GPT-5) to both generate and answer the QA pairs, it solves almost all questions correctly. 
We hypothesize that this is because constraining the MLLM to produce visually verifiable questions encourages it to generate relatively simple items whose answers lie well within GPT-5’s existing capabilities, thereby failing to probe the model’s unknown weaknesses or more challenging failure modes.

\item \textbf{General Video Question.} 
The introduction of visual information may lead the MLLM to generate overly general questions that can be answered without any sports-specific understanding, which is misaligned with our evaluation objective.
As shown in Fig.~\ref{supple:mllm_case3}, the generated question asks about the phase of the NBA season, which can be answered directly from the scoreboard in view 1.
However, this question is unrelated to sports knowledge and thus misaligned with our evaluation goal.

\end{itemize}

\begin{figure*}[t]
\centering
\includegraphics[width=1.0\linewidth]{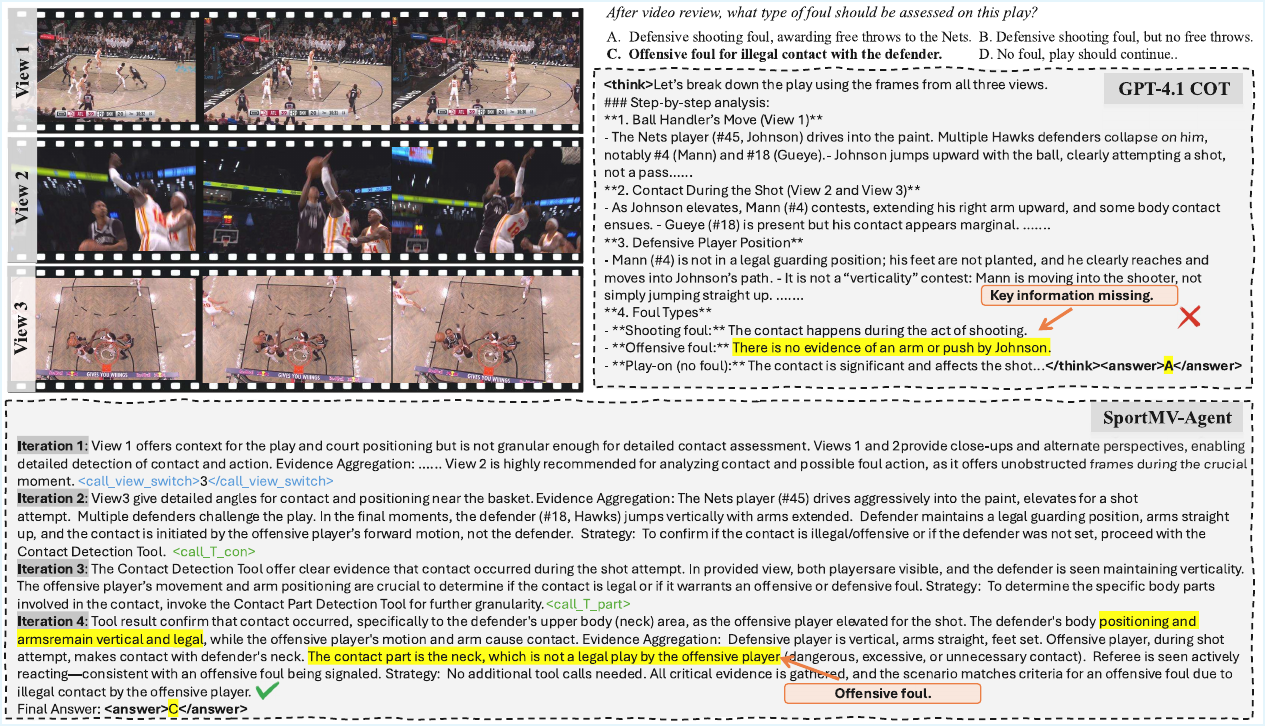}
\caption{
    \textbf{Qualitative example.} 
    Comparison between the reasoning outputs of \ouragent~and the COT baseline. The COT strategy suffers from visual missing: it fails to consider the critical evidence visible in \texttt{View 3}, incorrectly classifying the foul as a defensive shooting foul. In contrast, \ouragent~actively selects the most informative viewpoint (\texttt{View 3}), capturing the offensive player's illegal contact with the defender's neck. By synthesizing this with consistent observations of the defender’s legal verticality and positioning across other views, \ouragent accurately concludes an offensive foul.
}
\label{fig:supple_example}
\end{figure*}

\section{More Qualitative Example}
\label{example_supple}
In this section, we provide a qualitative comparison result (Fig.~\ref{fig:supple_example}) in addition between our SportMV-Agent and the COT strategy.
As illustrated in Fig.~\ref{fig:supple_example}, CoT fails to provide a correct ruling for this complex ADR case. A closer inspection of the CoT reasoning process reveals that its lack of dynamic viewpoint switching leads it to overlook the illegal contact initiated by the offensive player in View 3. In contrast, SportMV-Agent successfully identifies this infraction by strategically switching perspectives and invoking specialized contact-detection tools. By capturing the upper-body contact (specifically at the neck) and synthesizing information across multiple views, our method correctly classifies the incident as an offensive foul in accordance with official regulations.

\section{More Implementation Details}
As mentioned in the main paper, we leverage Qwen2.5-VL-72B~\cite{bai2025qwen2} as the action recognition tool, contact detection tool and contact part detection tool in our proposed agentic framework.
The output of the action recognition and contact part detection tool comprises the Top-3 retrieval results and their corresponding confidence scores.
In this section, we will provide more details.
For action recognition, we restrict the model's output space to the predefined action taxonomies of the respective datasets. Specifically, for soccer tasks, the model is constrained to the action classes defined in SoccerNet\cite{giancola2018soccernet}; for basketball tasks, it is limited to the categories within FineSports\cite{finesports}. This definition of action sets is comprehensive enough for the relevant sports categories, effectively preventing 'open-set' hallucinations and facilitating a more objective evaluation of the reasoning pipeline.
For the contact part detection tool, we employ a hierarchical classification strategy.
The model is required to first categorize the contact as either upper body or under body before specifying the exact anatomical part. This coarse-to-fine approach reduces initial search complexity and minimizes fine-grained confusion, mirroring the systematic observation process of professional referees.

\begin{table}[t]
  \centering
  \small
  \setlength{\tabcolsep}{6pt}
  \begin{tabular}{ll}
  \toprule
  Sport & Official video-review source \\
  \midrule
  Basketball          & NBA Replay Center \\
  Soccer              & Video Assistant Referee (VAR) \\
  Cricket             & Decision Review System (DRS) \\
  Rugby               & Television Match Official (TMO) \\
  American football   & Instant Replay review \\
  Ice hockey          & NHL Situation Room review \\
  Baseball            & MLB Replay Review \\
  Tennis              & Hawk-Eye challenge \\
  Volleyball          & Video Challenge System \\
  Australian football & AFL Score Review \\
  \bottomrule
  \end{tabular}
  \caption{Official video-review mechanism used as the multi-view source for each of the ten sports in \ourbench.}
  \label{tab:sources}
\end{table}
\begin{figure*}[t]
\centering
\includegraphics[width=1.0\linewidth]{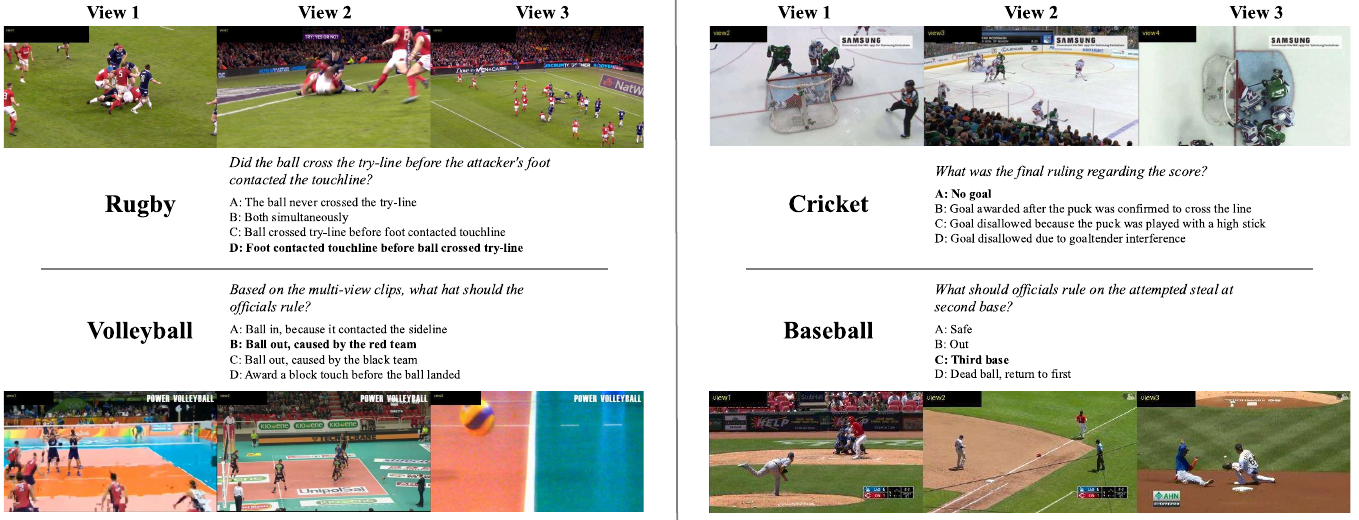}
\caption{\textbf{QA examples from \ourbench.} We show four examples drawn from baseball, volleyball, cricket, and rugby. Each example pairs a question and its four answer options with the corresponding multi-view clips.  Deriving the correct answer requires aggregating evidence across cameras rather than relying on a single view.}
\label{supple:example_qa}
\end{figure*}
\section{More Data Construction Details}
\subsection{Video Sources Collection}
\ourbench~is built from \emph{officiating-review} (replay) broadcasts. Our key observation is that, to adjudicate a contested play, every major sport operates an official video-review protocol that replays the same decisive moment from several broadcast cameras and announces an authoritative ruling. Such broadcasts are thus an ideal source of multi-view data: a single clip already contains the \emph{same event} from \emph{multiple angles}, together with a ground-truth decision that we later use as the answer source for QA construction (Stage~2). Tab.~\ref{tab:sources} lists the review mechanism we target for each of the ten sports. Basketball clips are collected from the NBA official replay archive, soccer clips are adopted from SoccerNet~\cite{giancola2018soccernet}, and the remaining eight sports are crawled from a public video platform (YouTube).

\noindent\textbf{Crawling procedure.} For each of the eight crawled sports, we design a set of \emph{event-level} search queries that target single-event review clips rather than long compilations, e.g., \textit{``cricket DRS review overturned out decision''}, \textit{``rugby TMO try or no try''}, and \textit{``tennis Hawk-Eye challenge ball in or out''}. For every query, candidate videos are retrieved and filtered on the fly by three complementary rules:
\begin{itemize}
    \item \textbf{Duration window.} We keep only clips whose length matches a typical single-event review, using a per-sport window of $15$--$45$\,s for fast reviews (cricket, rugby, tennis, Australian football) and $15$--$90$\,s for longer replay segments (American football, ice hockey, baseball, volleyball). This removes both trivially short junk and hour-long compilations, talks, or full matches.
    \item \textbf{Positive title requirement.} A candidate is retained only if its title matches a per-sport keyword whitelist covering the sport name and its officiating jargon (e.g. \texttt{DRS}, \texttt{TMO}, \texttt{VAR}, \texttt{Hawk-Eye}, \texttt{situation room}, \texttt{score review}). This is our primary anti-drift guard, preventing cross-sport contamination such as an American-football query returning soccer VAR clips.
    \item \textbf{Blocklist.} We reject titles matching an explainer/compilation blocklist (e.g., \textit{explainer}, \textit{top~10}, \textit{compilation}, \textit{reaction}, \textit{press conference}) and video-game footage (e.g., \textit{Madden}, \textit{2K}, \textit{FIFA}), which otherwise leak in when genuine in-sport review clips are scarce.
\end{itemize}
All retrieved videos are downloaded at up to 720p, deduplicated across runs via a global download archive, and capped per sport to maintain balance.
For each clip, we retain the official decision, together with its metadata (title, uploader, and description recording the verdict), as authoritative textual ground truth.
All material is used strictly for non-commercial academic research.

\subsection{Multi-View Segmentation and Filtering.}
Since review footage interleaves distinct viewpoints of the same action within a single continuous stream, we design a three-step procedure to recover a clean set of camera viewpoints. \textbf{(i) Shot segmentation.} We apply shot-boundary detection to partition each review sequence into consecutive shots. \textbf{(ii) Content filtering.} Broadcast streams are heavily interleaved with non-action content, such as scoreboards, graphics, decision cards, and tight close-ups of players or referees; treating every shot as a viewpoint therefore severely over-counts. With a single vision-language model (GPT-4.1) call per clip, we classify the representative frame of every shot as \emph{game}, \emph{close-up} or \emph{graphic}, and, for \emph{game} shots, additionally assign a coarse camera-viewpoint label (e.g., \textit{broadcase wide}, \textit{high behind goal}, \textit{sideline}). We retain only game shots and discard all graphic and close-up shots. \textbf{(iii) Viewpoint clustering.} As the broadcast repeatedly cuts back and forth between the same cameras, multiple shots may correspond to an identical viewpoint. We extract DINOv2 features from the retained shots and perform agglomerative clustering to merge visually redundant shots, yielding the set of genuinely distinct camera viewpoints. Only instances with at least two distinct viewpoints are kept as valid multi-view samples, while single-view instances are discarded. Notably, the resulting multi-view clips for a given event are not temporally synchronized, which further increases task difficulty by requiring the model to establish correspondence across unaligned views.

\subsection{Human Filtering and Refining}
The automatic pipeline (Stages~1--3) already removes most low-quality items, but a small fraction of QA pairs remain ambiguous, weakly grounded, or answerable without watching the videos. We therefore conduct a final round of expert human filtering and refinement.

\noindent\textbf{Annotators.}
We recruit five annotators with prior experience in the corresponding sports and their officiating rules. Before annotation, all annotators are given a unified guideline describing the three question types (PAR, REI, ADR), the definition of a valid multi-view question, and the reasons an item should be rejected.

\noindent\textbf{Review protocol.}
Each item is presented together with its question, four options, and the associated multi-view clips. An annotator first attempts to answer the question from the videos alone, and then judges the item against four criteria:
\begin{itemize}
    \item \textbf{Visual answerability.} The correct answer must be derivable from the multi-view clips, not from commonsense or textual priors.
    \item \textbf{Grounding and uniqueness.} The gold answer must be consistent with the official ruling, and exactly one option should be correct.
    \item \textbf{Clarity.} The question and options must be unambiguous and free of leaking cues.
\end{itemize}
For each item, an annotator either \emph{keeps}, \emph{edits}, or \emph{discards} it.
Editing includes rephrasing the question for neutrality and clarity, and revising distractors so that they remain plausible yet clearly incorrect.
To ensure reliability, every item is reviewed by at least two annotators, and an item is retained only when they agree; disagreements are resolved by a third annotator.
This stage removes a further \textbf{13.81\%} of the candidate items and refines the wording of many others, yielding the final \qanumber~QA pairs.
The resulting corpus exhibits high linguistic quality and multimodal alignment, providing a reliable basis for evaluating multi-view, fine-grained visual-textual understanding in sports officiating.

\section{Dataset QA Examples}
To provide a more intuitive understanding of \ourbench, Fig.~\ref{supple:example_qa} presents representative QA examples in different sports.
Each example pairs a question and four answer options (with the correct answer highlighted) with its multi-view clips. Answering correctly requires aggregating evidence across cameras rather than relying on a single view.

\end{document}